# BPF Algorithms for Multiple Source-Translation Computed Tomography Reconstruction

Zhisheng Wang[1,2,#], Haijun Yu[3,#], Yixing Huang[4], Shunli Wang[1,2], Song Ni[3], Zongfeng Li[1,2], Fenglin Liu[3,*], Junning Cui[1,2,*]

1. *Center of Ultra-Precision Optoelectronic Instrument Engineering, Harbin Institute of Technology, Harbin 150080, China*
2. *Key Lab of Ultra-Precision Intelligent Instrumentation (Harbin Institute of Technology), Ministry of Industry and Information Technology, Harbin 150080, China*
3. *Key Laboratory of Optoelectronic Technology and Systems, Ministry of Education, Chongqing University, Chongqing 400044, China*
4. *Oncology, University Hospital Erlangen, Friedrich-Alexander-University Erlangen-Nuremberg, 91054 Erlangen, Germany*

*These authors contributed equally to this work: Zhisheng Wang, Haijun Yu*

*Corresponding authors: Junning Cui, Fenglin Liu*

**Email:** cuijunning@126.com, liufl@cqu.edu.cn

**ABSTRACT**: Micro-computed tomography (micro-CT) is a widely used state-of-the-art instrument employed to study the morphological structures of objects in various fields. However, its small field-of-view (FOV) cannot meet the pressing demand for imaging relatively large objects at high spatial resolutions. Recently, we devised a novel scanning mode called multiple source translation CT (mSTCT) that effectively enlarges the FOV of the micro-CT and correspondingly developed a virtual projection-based filtered backprojection (V-FBP) algorithm for reconstruction. Although V-FBP skillfully solves the truncation problem in mSTCT, it requires densely sampled projections to arrive at high-resolution reconstruction, which reduces imaging efficiency. In this paper, we developed two backprojection-filtration (BPF)-based algorithms for mSTCT—S-BPF (derivatives along source) and D-BPF (derivatives along detector). D-BPF can achieve high-resolution reconstruction with fewer projections than V-FBP and S-BPF. Through simulated and real experiments conducted in this paper, we demonstrate that D-BPF can reduce source sampling by 75% compared with V-FBP at the same spatial resolution, which makes mSTCT more feasible in practice. Meanwhile, S-BPF can yield more stable results than D-BPF, which is similar to V-FBP.

## 1. Introduction

In recent years, micro-computed tomography (micro-CT) has been widely used in various fields to conduct a non-invasive and high-resolution investigation of the objects' internal microstructure [1–3]. Most commercial micro-CT systems employ a circular scanning mode to acquire the projection data from various direction angles within 360 degrees. During their scan process, the micro-focus X-ray source and the flat-panel detector (FPD) are stationary, the high-precision air-bearing rotation stage rotating the detected object, which is abbreviated as RCT scanning mode in this paper. In the RCT system, the field-of-view (FOV) is the inscribed circle of the triangle formed by the detector and X-ray source, which is further mainly determined by the detector size, especially for the high magnification ratio. The object must be restricted within the FOV to avoid the truncation artifacts in the reconstruction images. Commercially available FPDs are only capable of providing a FOV ranging from 0.3 mm to 3.0 mm with a spatial resolution between 0.1 μm and 1.0 μm. However, in many scenarios, a larger FOV for high resolution is demanded, such as imaging the biological specimens or fossils that cannot be cropped into an appropriate size [4], thereby characterizing the 3D distribution of the tiny features over a large area [5, 6].

Over the past decades, several techniques have been developed to enlarge the FOV of micro-CT, including the detector offset [7–10], traverse-continuous-rotate scanning [11], rotation-translation-translation (RTT) multi-scan mode [12], rotation-translation (RT) multi-scan mode [13], elliptical trajectory [14], complementary circular scanning [15], and rotated detector [16]. Very recently, we developed a multiple source translation method to enlarge the FOV of the micro-CT system, which is mainly determined by the adjustable source translation distance. The multiple source translation CT (mSTCT) consists of multiple STCTs with different translation angles. In each STCT, the object is placed close to the source but far from the FPD to achieve a large geometry magnification [17]. During the scanning, the source translates along a line parallel to the fixed FPD to acquire projection data from different angles.

In mSTCT, the projection views are truncated since the cone-beam X-rays from each source position can only illuminate a portion of the object. It poses a challenge to apply the filtered backprojection (FBP) algorithm to reconstruct the image, which needs non-truncated projection views in the filtering step. Over the decades, different methods have been proposed to tackle the truncated projection views in enlarging FOV scanning modes, which can be summarized as the following categories: weighted projection-based algorithms [18], virtual projection-based algorithms [19, 20], backprojection filtration (BPF) algorithms [21–24], iterative algorithms [25–27], and deep learning methods [28-32]. Wherein, the weighted projection-based algorithms leverage the data redundancy in the truncated region to suppress truncation artifacts by applying a proper weighting function before or after the filtering step, which has been successfully applied in the detector offset scanning mode. The virtual projection-based algorithms convert the truncated projection views into a set of virtual projection views of non-truncated, which effectively tackle truncation artifacts in the RTT and RT multi-scan modes. The BPF algorithms utilize the local property of the differential operator to mitigate the error caused by truncated projection views, which have been successfully applied to the rotation-translation-translation (RTT) multi-scan and the detector offset scan modes.

In our previous work, we developed a virtual projection-based FBP (V-FBP) algorithm for mSTCT, where the virtual projection view is a set of measured rays divergent from each detector element [19]. Benefiting from the geometry of mSTCT, the acquired truncated projection views can be regrouped into the virtual projection views without interpolation. In V-FBP, each virtual projection view is filtered individually, which is equivalent to filtering along the source trajectory, whose cutoff frequency depends on the source sampling frequency. The Nyquist criterion states that the source sampling frequency should be greater than double of the highest frequency in projection data to avoid loss of high-frequency information and spatial resolution. Consequently, we need to sample thousands of projection views to achieve high-resolution reconstruction via V-FBP, leading to a time-consuming acquisition process and huge memory requirements to store the dataset.

In this paper, we develop a BPF algorithm for mSTCT to address the truncated projection views at low source sampling frequencies. BPF consists of two steps: calculating the differentiated backprojection (DBP) image by back-projecting the derivative of the projections according to the scanning mode; and obtaining the final reconstruction by calculating the finite inversion of Hilbert transform on the DBP image. Here, the derivative is implemented along the detector, thereby relaxing the requirement for source sampling. To avoid the derivative errors at the truncated boundary propagating to the whole DBP image, we introduce a trick that throws away the derivative data at the truncation boundary.

Besides, implementing the derivative along the source trajectory, we get the other DBP formula. For clarification, we denote the DBP image from the derivative along the detector as D-DBP and along the source as S-DBP. We observe that these two formulae result in distinct reconstruction results under different projection views, and we have theoretically analyzed this phenomenon in this paper.

Finally, we derive the D-DBP and S-DBP formulae in both the two-dimensional (2D) and three-dimensional (3D) cases for mSTCT scanning mode. We conduct the simulated and real experiments to demonstrate the efficacy of the developed BPF algorithm for mSTCT reconstruction to tackle the truncated projection views while reducing the number of source samplings.

This paper is organized as follows. In Section 2, we review the scanning geometry of mSTCT, point out the problem of truncated projections, and then introduce the 2D parallel BPF algorithm for the circular trajectory. In Section 3, two types of DBP formulae for STCT are derived in detail, the finite Hilbert inverse per STCT is proposed, and two types of 2D and 3D BPF-type algorithms are built. Section 4 describes the simulated and realistic experiment setup, which is followed by experimental results in Section 5. In Section 6 and Section 7, the discussion and conclusion end the paper.

## 2. Theory

*2.1 mSTCT scanning mode and truncated projections*

As illustrated in Fig. 1, mSTCT consists of multiple STCTs with different translation angles. In each STCT, the object is positioned close to the X-ray source, while the source is then translated parallel to the fixed FPD, enabling the acquisition of projections from various angles. Mathematically, the projection from a STCT with translation angle $\theta$ can be formulated as

$$p_\theta(\lambda, u) = \int_{-\infty}^{+\infty}\int_{-\infty}^{+\infty} f(x,y)\delta(x\cos\varphi + y\sin\varphi - r)\,dxdy \tag{1}$$

with

$$r = \frac{\lambda h + ul}{\sqrt{(l+h)^2 + (\lambda-u)^2}}, \quad \varphi = \theta + \arctan(\frac{\lambda-u}{l+h}). \tag{2}$$

Here, $l$ denotes the isocenter-to-source trajectory distance; $h$ represents the isocenter-to-detector distance; $\lambda \in [-s, s]$ is the source index, where $s$ is the half length of the source trajectory; $u \in [-d, d]$ is the detector index, where $d$ is the half length of the detector; $r$ is the signed distance from the origin to the measured ray; and $\varphi$ is the angle between the ray and the positive $y$-axis, as shown in Fig. 1(c). The FOV of mSTCT is defined as the region with complete angular coverage [17], whose radius is

$$R_1 = \frac{sh - dl}{\sqrt{(l+h)^2 + (s+d)^2}}. \tag{3}$$

Fig. 1(d) illustrates the projection from STCT. We observe that the projection along the detector (axis-$u$) is truncated, posing a challenge to reconstruction using FBP-type algorithms since the filtering step requires non-truncated projection data. Truncated boundaries introduce infinite high-frequency components in the frequency space and produce the Gibbs phenomenon after the filtering step, finally leading to streaking artifacts in reconstructions. Based on the geometry of STCT, we find that the projection along the source trajectory (axis-$\lambda$) is non-truncated. However, if a filter is implemented along the source, we need to sample thousands of projections to guarantee that the cut frequency of the filter is larger than the high-frequency components. Therefore, in this paper, we try to leverage the local property of the derivative operator in BPF algorithm to address the truncated projection.

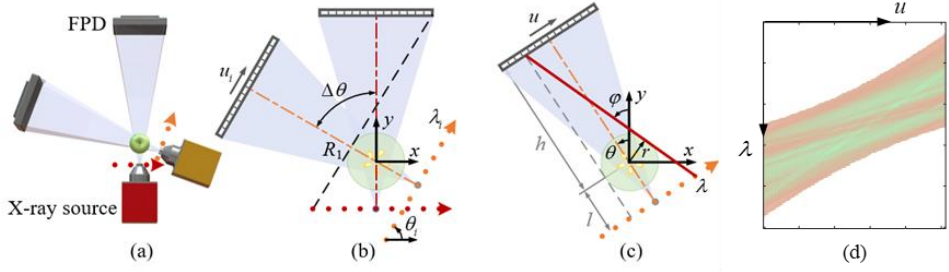

**Fig. 1** mSTCT scanning mode. (a) 3D geometry of mSTCT with two STCTs; (b) 2D transverse section of (a); (c) 2D geometry of one STCT with arbitrary translation angle; (d) Projection data from one STCT.

*2.2 2D parallel BPF algorithm*

Noo et al. [22] derived an explicit 2D parallel DBP formula as

$$b_\eta(\vec{x}) = \int_0^\pi \text{sgn}(\sin(\varphi-\eta)) \frac{\partial}{\partial r} \bar{p}(\varphi,r) \text{d}\varphi, \qquad (4)$$

where $\vec{x}=(x,y)$ is the reconstructed point; $\text{sgn}(\cdot)$ represents the signum function; $\bar{p}(\varphi,r)$ denotes parallel-beam projection based on the relation in Eq. (1). $b_\eta(\vec{x})$ is the DBP image, which has the following relation with the original image

$$b_\eta(\vec{x}) = -2\pi \mathcal{H}_\eta f(\vec{x}), \qquad (5)$$

where $\mathcal{H}_\eta f(\vec{x})$ denotes the Hilbert transform of $f(\vec{x})$ along lines at angle $\eta$, as illustrated in Fig. 2.

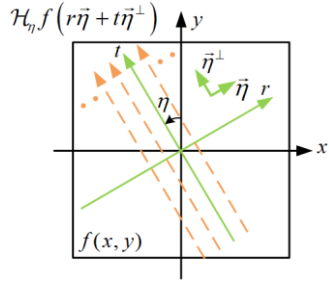

**Fig. 2** Description of the coordinate systems in Hilbert transform.

Finally, $f(\vec{x})$ can be recovered by using the finite inverse formula of Hilbert transform. Several inversion formulae have been developed [33, 34]. The one proposed by Mikhlim [35] will be applied here.

**Definition 1**. *If $\mathcal{H}_\eta f(r\vec{\eta}+t\vec{\eta}^\perp)$ is known for all $t \in [L_r, U_r]$, its finite inversion of Hilbert transform $\mathcal{H}_\eta^{-1}$ is*

$$\mathcal{H}_\eta^{-1}\left(\mathcal{H}_\eta f(r\vec{\eta}+t\vec{\eta}^\perp)\right) = f(r\vec{\eta}+t\vec{\eta}^\perp) = \frac{-1}{\sqrt{(t-L_r)(U_r-t)}} \left( \int_{L_r}^{U_r} \sqrt{(t'-L_r)(U_r-t')} \frac{\mathcal{H}_\eta f(r\vec{\eta}+t'\vec{\eta}^\perp)}{\pi(t-t')} \text{d}t' + C_r \right). \qquad (6)$$

*If $f(r\vec{\eta}+t\vec{\eta}^\perp)$ is known to be zero for $t \notin [L_r+\varepsilon_r, U_r-\varepsilon_r]$, $C_r$ can be determined by*

$$C_r = -\int_{L_r}^{U_r} \sqrt{(t'-L_r)(U_r-t')} \frac{\mathcal{H}_\eta f(r\vec{\eta}+t'\vec{\eta}^\perp)}{\pi(t_0-t')} \text{d}t'. \qquad (7)$$

Here, $\vec{\eta} = (\cos\eta, \sin\eta)$, $\vec{\eta}^{\perp} = (-\sin\eta, \cos\eta)$, and $\vec{x} = r\vec{\eta} + t\vec{\eta}^{\perp}$. $\varepsilon_r$ is a small positive.

## 3. Methods

### 3.1 DBP formulae for mSTCT

Because the inversion of Hilbert transform is independent of CT scan mode, we only need to derive the DBP formula to achieve BPF reconstruction for mSTCT. Like the form of the FBP-type algorithm, we can describe the derivative in Eq. (4) using a filtering kernel $\delta'(r)$, which is the derivative of the Dirac delta function $\delta(r)$ and has the following properties:

**Property 1:** $\int_{-\infty}^{\infty} f(r')\delta'(r-r')\mathrm{d}r' = f'(r)$,

**Property 2:** $\delta'(ar) = \delta'(r)/a^2$ for $a > 0$.

Then, we can rewrite Eq. (4) as

$$b_{\eta}(\vec{x}) = \frac{1}{2}\int_0^{2\pi}\int_{-\infty}^{\infty} \mathrm{sgn}(\sin(\varphi-\eta))\bar{p}(\varphi,r')\delta'(x\cos\varphi + y\sin\varphi - r')\mathrm{d}r'\mathrm{d}\varphi. \tag{8}$$

Replacing the parallel-beam projection $p(\varphi,r)$ with the fan-beam projection $p_{\theta}(\lambda,u)$ according to Eq.(1), we can derive an initial fan-beam DBP formula for STCT as (*details in Supplementary A*)

$$b_{\eta}^{\theta}(\vec{x}) = \frac{1}{2}\int_{-\infty}^{+\infty}\int_{-\infty}^{+\infty} \mathrm{sgn}\left[\sin\left(\theta + \arctan\left(\tfrac{\lambda-u}{l+h}\right) - \eta\right)\right] \frac{(l+h)^2}{\left[(l+h)^2 + (\lambda-u)^2\right]^{\frac{3}{2}}} p_{\theta}(\lambda,u)$$
$$\times \delta'\left(\frac{(l+h)(x\cos\theta + y\sin\theta) - uL - \lambda H}{\sqrt{(l+h)^2 + (\lambda-u)^2}}\right)\mathrm{d}\lambda\mathrm{d}u, \tag{9}$$

Here, $H = x\sin\theta - y\cos\theta + h$ is the perpendicular distance from point $\vec{x}$ to the detector; $L = -x\sin\theta + y\cos\theta + l$ is the perpendicular distance from point $\vec{x}$ to the source trajectory.

In Eq. (9), we can see that the filtering kernel $\delta'(\cdot)$ can either take partial derivatives of $u$ or $\lambda$. For this reason, our insight is that it can derive two different forms of reconstruction algorithms, which may produce different interesting reconstruction results.

### (1) D-DBP formula

We reformulate the filtering kernel in Eq. (9) as

$$\delta'\left(\frac{L}{\sqrt{(l+h)^2 + (\lambda-u)^2}}\left[\frac{(l+h)(x\cos\theta + y\sin\theta) - \lambda H}{L} - u\right]\right). \tag{10}$$

According to its **Property 1**, Eq. (9) can be rewritten as

$$Db_{\eta}^{\theta}(\vec{x}) = \frac{1}{2}\int_{-\infty}^{+\infty}\frac{1}{L^2}\mathrm{d}\lambda\int_{-\infty}^{+\infty} \mathrm{sgn}\left[\sin\left(\theta + \arctan\left(\tfrac{\lambda-u}{l+h}\right) - \eta\right)\right] \frac{(l+h)^2}{\sqrt{(l+h)^2 + (\lambda-u)^2}} p_{\theta}(\lambda,u)\delta'(u^* - u)\mathrm{d}u, \tag{11}$$

with

$$u^* = \frac{(l+h)(x\cos\theta + y\sin\theta) - \lambda H}{L}, \tag{12}$$

which denotes the location of the detector unit whose X-ray passes through the point $\vec{x}$ and source position $\lambda$. Based on the **Property 2** of the filtering kernel, Eq. (11) is equivalent to

$$Db_\eta^\theta(\vec{x}) = \frac{1}{2}\int_{-\infty}^{+\infty}\frac{1}{L^2}\frac{\partial}{\partial u}\left\{\text{sgn}\left[\sin\left(\theta+\arctan\left(\tfrac{\lambda-u}{l+h}\right)-\eta\right)\right]\frac{(l+h)^2}{\sqrt{(l+h)^2+(\lambda-u)^2}}p_\theta(\lambda,u)\right\}\Bigg|_{u=u^*}d\lambda. \tag{13}$$

However, due to the discontinuity caused by the signum function in the argument of the derivative, Eq. (13) is difficult to implement accurately [22]. This problem can be avoided by applying the product rule (*details in Supplementary B*). Therefore, we obtain a more implementable formula

$$Db_\eta^\theta(\vec{x}) = \frac{1}{2}\int_{-\infty}^{+\infty}\frac{1}{L^2}\text{sgn}\left[\sin\left(\theta+\arctan\left(\tfrac{\lambda-u^*}{l+h}\right)-\eta\right)\right]\frac{\partial}{\partial u}\left\{\frac{(l+h)^2}{\sqrt{(l+h)^2+(\lambda-u)^2}}p_\theta(\lambda,u)\right\}\Bigg|_{u=u^*}d\lambda + \text{S.T.}, \tag{14}$$

with

$$\text{S.T.} = \begin{cases} \dfrac{1}{2|L|}\dfrac{(l+h)p_\theta(\lambda_1,u_1^*)}{\sqrt{(l+h)^2+(\lambda_1-u_1^*)^2}}, & -\pi/2 < \theta-\eta < \pi/2 \\ 0, & \theta-\eta \in (-\infty,-\pi/2]\cup[\pi/2,\infty) \end{cases}, \tag{15}$$

where $\lambda_1 = x\cos\theta + y\sin\theta - L\tan(\theta-\eta)$ and $u_1^* = x\cos\theta + y\sin\theta + H\tan(\theta-\eta)$.

When $\theta-\eta = \pi/2$, we have S.T. = 0 and $\text{sgn}\left[\sin\left(\theta+\arctan\left(\tfrac{\lambda^*-u}{l+h}\right)-\eta\right)\right] \equiv 1$. Eq. (14) becomes a more succinct and refined D-DBP formula for STCT

$$Db_\eta^\theta(\vec{x}) = \frac{1}{2}\int_{-s}^{+s}\frac{1}{L^2}\frac{\partial}{\partial u}\left\{\frac{(l+h)^2}{\sqrt{(l+h)^2+(\lambda-u)^2}}p_\theta(\lambda,u)\right\}\Bigg|_{u=u^*}d\lambda. \tag{16}$$

In fact, differentiation is a local operator, which can prevent global operators like slope filters from spreading truncation errors to the entire filtering line. However, the differentiation along the detector still produces significant errors at the truncation points, manifested as a sharp value in the differentiated data that cannot be removed. To mitigate this error, our method is to add the closest dataset to the projection edge.

**Tip 1**: *add the closest set of data to the projection edge.*

*(2) S-DBP formula*

If we reformulate the filtering kernel in Eq. (9) as

$$\delta'\left(\frac{H}{\sqrt{(l+h)^2+(\lambda-u)^2}}\left[\frac{(l+h)(x\cos\theta+y\sin\theta)-uL}{H}-\lambda\right]\right), \tag{17}$$

we obtain the other DBP formula

$$Sb_\eta^\theta(\vec{x}) = \frac{1}{2}\int_{-d}^{+d}\frac{1}{H^2}\frac{\partial}{\partial\lambda}\left\{\frac{(l+h)^2}{\sqrt{(l+h)^2+(\lambda-u)^2}}p_\theta(\lambda,u)\right\}\Bigg|_{\lambda=\lambda^*}du, \tag{18}$$

with

$$\lambda^* = \frac{(l+h)(x\cos\theta + y\sin\theta) - uL}{H}, \quad (19)$$

which denotes the location of the source position, whose X-ray passes through the point $\vec{x}$ and detector unit $u$.

Comparing with Eq. (16), Eq. (18) fundamentally differs in that:

(1) Derivatives are implemented in different directions. Based on the **Property 2** of the filtering kernel, the derivative is equivalent to the filtering process. In D-DBP, the cut-frequency of the filtering is determined by the detector sampling interval, while it depends on the source sampling interval in S-DBP. Therefore, these two formulae can produce different resolutions in reconstruction images.

(2) The geometry weights in backprojection are different. The geometry weight is $1/L^2$ in D-DBP and $1/H^2$ in S-DBP. Since $H$ is always larger than $L$ for high-resolution imaging, S-DBP will produce more stable results than D-DBP when noise is present.

(3) In STCT, the projection is truncated along the detector direction but non-truncated along the source direction. Therefore, compared with D-DBP, S-DBP can totally avoid the errors caused by the truncated projections.

### *3.2 BPFs for mSTCT*

After obtaining DBP images via Eq. (16) or Eq. (18), we can get the reconstruction image $f_\theta(\vec{x})$ based on Eqs. (5)–(7) for each STCT. We denote the BPF reconstructions from D-DBP and S-DBP images as D-BPF and S-BPF, respectively, which are:

(1) **D-BPF** reconstruction

$$f_\theta(\vec{x}) = -\mathcal{H}_\eta^{-1}\left(Db_\eta^\theta(x)/2\pi\right), \quad (20)$$

(2) **S-BPF** reconstruction

$$f_\theta(\vec{x}) = -\mathcal{H}_\eta^{-1}\left(Sb_\eta^\theta(x)/2\pi\right), \quad (21)$$

Finally, because of the linearity of the BPF algorithm, the final reconstruction image for mSTCT can be calculated via

$$f(\vec{x}) = \sum_{i=1}^{T} f_{\theta_i}(\vec{x}). \quad (22)$$

Nevertheless, practical implementation of the BPF algorithm for mSTCT still faces certain challenges, including:

(1) In mSTCT, some projections have been sampled twice, which need to be weighted to avoid artifacts.
(2) In Eqs. (16) or (18), the Hilbert transform direction is $\eta = \theta - \pi/2$, which is intractable to implement in discrete images. However, the finite Hilbert inverse can be easily implemented if $\eta = k\pi/2, \ (k=0,1,2\cdots)$.
(3) The DBP image within the finite range $r$ and all $t \in [L_r, U_r]$ is divided by $-2\pi$ (see Eq. (5)) and used to complete the Hilbert transform. To improve the reconstruction quality, the DBP image needs to be obtained on a finite matrix larger than the preset reconstructed area [22]. Moreover, due to the limited angle scanning of one STCT, X-shaped finite angle artifacts may be formed in any direction under mSTCT. As a result, identifying regions with known zeros in the DBP images to calculate $C_r$ (see Eqs. (6) and (7)) becomes challenging and inefficient.

To address these challenges, we developed following corresponding techniques.

**Technique 1.** *In Eq. (16) or (18), before the derivation, the projection should multiple a redundancy weighting function, similar to our previous work* [19].

**Technique 2.** *Denote $\mathcal{R}_\eta$ as a rotation transform*

$$f(r,t) = \mathcal{R}_\eta^{-1}(f(x,y)), \quad \text{s.t. } r = x\cos\eta + y\sin\eta; \ t = -x\sin\eta + y\cos\eta. \tag{23}$$

*Then, finite inversion of Hilbert transform with an arbitrary angle can be easily implemented according to $\mathcal{H}_\eta^{-1} = \mathcal{R}_\eta \mathcal{H}_0^{-1} \mathcal{R}_{-\eta}(f(x,y))$.*

**Technique 3**: *Complete the backprojection in a zero-filled extended matrix (larger than the reconstructed matrix). With the aid of reconstruction over larger areas and the rotation operation in technique 2, the DBP images with arbitrary X-shaped finite angle artifacts will be flattened horizontally, and the resulting immediate image will have many rows with values of all zeros in the upper and lower regions, thereby simply finding enough of these rows to accurately calculate $C_r$.*

Overall, Table 1 illustrates the pseudo code of the finite Hilbert inverse function per STCT.

**Table 1.** The pseudo-code of finite Hilbert inverse function per STCT.

| Finite Hilbert inverse function per STCT |
|---|
| 1. **Input:** DBP image of one STCT $b_\eta^\theta(\vec{x})$; reconstructed size $I \times I$; number of padding zeros $p_0$; angle of Hilbert transform $\eta$. |
| 2. $\mathcal{H}_0 f_\theta(\vec{x}) \leftarrow \mathcal{R}_{-\eta}(b_\eta^\theta(\vec{x})/(-2\pi))$; |
| 3. Finding the row indexes $\vec{i}$ in $\mathcal{H}_\eta f_\theta(\vec{x})$ where all elements in the row are zeros; |
| 4. $\mathcal{R}_{-\eta} f_\theta(\vec{x}) \leftarrow \mathcal{H}_0^{-1}(\mathcal{H}_0 f_\theta(\vec{x}) \cdot \sqrt{(t'-L_r)(U_r-t')})$; |
| 5. $C_{x_1} \leftarrow \mathcal{M}(\mathcal{R}_{-\eta} f_\theta(\vec{i},:))$; |
| 6. $\mathcal{R}_{-\eta} f_\theta(\vec{x}) \leftarrow (\mathcal{R}_{-\eta} f_\theta(\vec{x}) - C_{x_1})/\sqrt{(t'-L_r)(U_r-t')}$; |
| 7. $f_\theta(\vec{x}) \leftarrow \mathcal{R}_\eta(\mathcal{R}_{-\eta} f_\theta(\vec{x}))$; |
| 8. $f_\theta(\vec{x}) \leftarrow f_\theta(1+p_0:p_0+I, 1+p_0:p_0+I)$; |
| 9. **Output:** The reconstructed result of one STCT $f_\theta(\vec{x})$. |

Table note: $\mathcal{M}(\cdot)$ is the operator used to average a matrix along the column direction.

Fig. 3 presents the flowchart of 2D BPF reconstructions for mSTCT. The figures in the first row describe the steps of processing the raw projection per STCT to obtain the DBP image. The figures in the second row illustrate the effectiveness of the finite Hilbert inverse per STCT. Note that when the angle relationship $\eta_i = \theta_i - \pi/2$ is set and the finite Hilbert inverse is performed along the source translation per STCT, the artifacts in the final reconstructed image are almost invisible. Besides, other key operations of the **Techniques** can be found in Fig. 3.

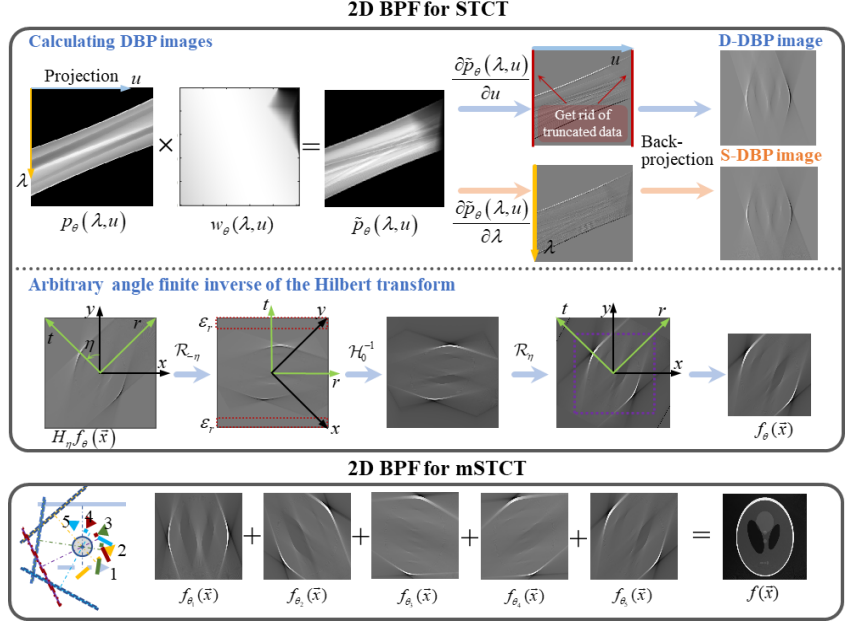

**Fig. 3** Flowchart of 2D BPF reconstructions for mSTCT.

*3.3 3D BPF reconstructions for mSTCT*

Similar to the 2D BPF reconstruction for mSTCT, we can first obtain the 3D BPF reconstruction for each STCT separately and then add them together to achieve 3D reconstructions for mSTCT. For 3D BPF reconstruction in each STCT, as shown in Fig. 4, we need to calculate the 3D DBP image along the line at an angle $\eta$ and then implement the inversion of Hilbert transform layer by layer to obtain the reconstruction image.

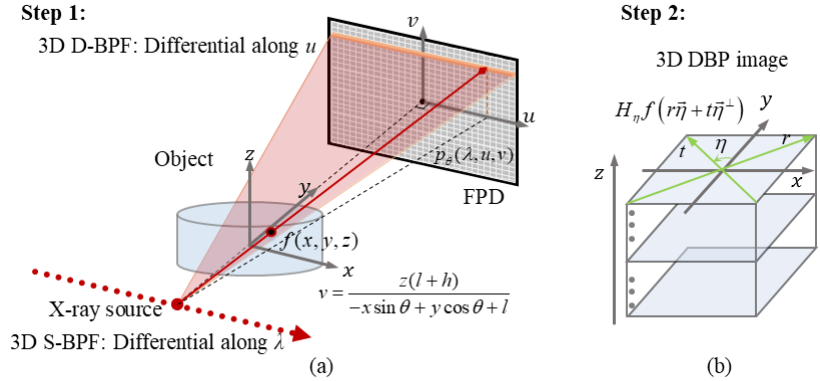

**Fig. 4** Schematic of 3D BPF reconstructions in each STCT. (a) Derivation in different directions. (b) Inversion of Hilbert transform layer by layer to obtain the reconstruction image.

The 3D DBP image can be calculated in a way similar to that in Feldkamp-David-Kress (FDK) algorithm [36]. Besides, if we implement derivatives in different ways, 3D DBP images also have two different formulae in STCT as

(1) D-DBP: Derivative along the detector ($u$-axis)

$$Db_\eta^\theta(\vec{x}) = \frac{1}{2}\int_{-s}^{+s} \frac{1}{L^2} \frac{\partial}{\partial u} \left\{ \frac{(l+h)^2 \cdot w_\theta(\lambda,u,v)}{\sqrt{(l+h)^2 + (\lambda-u)^2 + v^2}} p_\theta(\lambda,u,v) \right\} \Bigg|_{u=u^*} d\lambda, \qquad (24)$$

(2) S-DBP: Derivative along the source ($\lambda$-axis)

$$Sb_\eta^\theta(\vec{x}) = \frac{1}{2}\int_{-d}^{+d}\frac{1}{H^2}\frac{\partial}{\partial\lambda}\left\{\frac{(l+h)^2 \cdot w_\theta(\lambda,u,v)}{\sqrt{(l+h)^2+(\lambda-u)^2+v^2}}p_\theta(\lambda,u,v)\right\}\Bigg|_{\lambda=\lambda^*} du, \quad (25)$$

with $v = z(l+h)/L$, as shown in Fig. 4(a). Here, $\vec{x}=(x,y,z)$ is the reconstruction voxel; $p_\theta(\lambda,u,v)$ denotes the 2D projection taken at the source position $\lambda$; $w_\theta(\lambda,u,v)$ is the redundancy weighting function to deal with the redundant projection.

Fig. 5 describes the flowchart for 3D BPF for mSTCT, with the following steps: first, weighting the cone-beam projections of the $i$-th STCT ($i = 1, 2, ..., T$, and $T = 5$), where we apply the redundancy weights in the mid-plane to all rows of the FPD (i.e., $w_{\theta_i}(\lambda,u,v) = w_{\theta_i}(\lambda,u,0)$); second, in 3D D-BPF and S-BPF, derivating the weighted projections along the $u$-axis and $\lambda$-axis to obtain the derivative projections; third, backprojecting the derivative projections, then weighting $1/L^2$ and $1/H^2$ to get the 3D DBP images (see Eqs. (24) and (25), that is, $Db_{\theta_i}^{\eta_i}(\vec{x})$ and $Sb_{\theta_i}^{\eta_i}(\vec{x})$), respectively; fourth, reconstructing the incomplete 3D image of the $i$-th STCT ($f_{\theta_i}(\vec{x})$) by performing the finite Hilbert inverse layer by layer along the $z$-axis for the 3D DBP image; fifth, adding up all reconstructed results to obtain a complete volume ($f(\vec{x})$) by above steps.

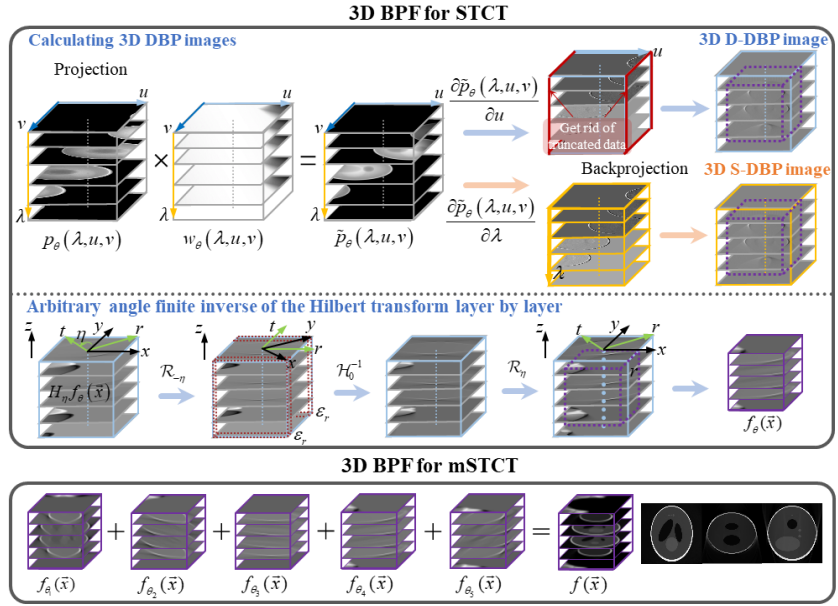

**Fig. 5** Flowchart of 3D BPF reconstructions for mSTCT.

## 4. Experiment Setup

In the experiments, our programs are implemented in Python and executed on a computer workstation equipped with an Intel(R) Xeon(R) Silver 4210R CPU running at 2.40 GHz and an NVIDIA GT730 GPU. We utilize the GPU-based ASTRA toolbox for efficient handling of the projection and backprojection steps.

*4.1 Simulated experiments*

To assess the performance of BPF algorithms for mSTCT, we conduct a series of quantitative numerical experiments using the parameters outlined in Table 2. For the 2D mSTCT geometry, we generate simulated projections by employing the FORBILD phantom with dimensions of 8.4 mm × 8.4 mm and 512 × 512 pixels. As a benchmark for comparison, we utilize the V-FBP algorithm. Furthermore, for the 3D mSTCT geometry, we generate simulated cone-beam projections by utilizing the 3D Shepp-Logan phantom with dimensions of 8.4 mm × 8.4 mm × 8.4 mm and 512 × 512 × 512 pixels. As a reference, we employ V-FDK, which is an approximate cone-beam reconstruction algorithm derived from V-FBP.

We utilize three metrics to assess the perceived quality of reconstructed images: root mean square error (RMSE), peak signal-to-noise ratio (PSNR), and feature similarity index measurement (FSIM). RMSE is employed to quantify the overall error between the reconstructed image and the reference phantom. PSNR is utilized to measure the noise level presented in the reconstructed image. FSIM is used to evaluate perceptual consistency, with its definition provided in reference [37]. A lower RMSE value, along with higher PSNR and FSIM values, indicates better image quality.

Table 2. Parameters for numerical experiments.

| Parameters | Values |
|---|---|
| Size of detector unit (mm$^2$) | 0.127×0.127 |
| Number of detector units $M \times M$ | 1024×1024 |
| Source-to-isocenter distance $l$ (mm) | 15 |
| Detector-to-isocenter distance $h$ (mm) | 190 |
| Source translation distance $2s$ (mm) | 20 |
| Number of STCTs $T$ | 5 |
| Interval translation angle $\Delta\theta$ (°) | 36.5 |
| Magnification ratio $k$ | 13.7 |
| Radius of FOV $R_1$ (mm) | 4.2 |
| Number of projections per STCT $N$ | 251,501,1001,2001 |

*4.2 Real experiments*

To validate the performance of BPF reconstructions for mSTCT using real data, we employ a micro-CT system depicted in Fig. 6. The system comprises a micro-focus X-ray source (L10321, Hamamatsu, Japan), a rotation platform (RGV100BL, Newport, USA), a FPD (PaxScan1313DX, Varian, USA), a translation platform (M-ILS250PP, Newport, USA), and a control system (Newport). For source translation, the source is mounted on a translation platform parallel to the fixed FPD. A bamboo specimen with a diameter of approximately 10 mm is affixed to the rotation platform for imaging purposes. The X-ray source operates at a tube voltage of 60 kV and a tube current of 70 μA. The geometry parameters utilized in the real experiments correspond to those employed in the simulated experiments outlined in Table 2.

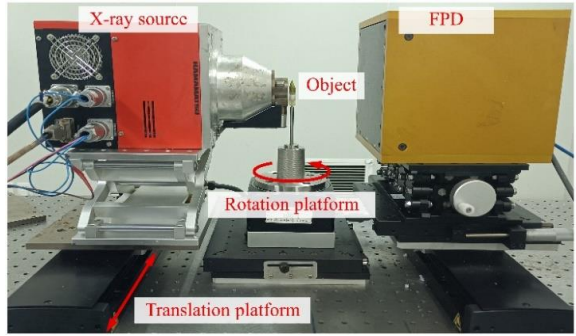

**Fig. 6** Experimental platform for mSTCT.

## 5. Results

*5.1 Results of simulated data*

*5.1.1 2D BPF reconstructions for mSTCT*

Fig. 7 presents the reconstructed images and their corresponding zoomed-in versions obtained using V-FBP, S-BPF, and D-BPF with varying numbers of projections. As the number of projections per STCT ($N$) increases, both V-FBP and S-BPF exhibit a gradual improvement in image sharpness, reducing blurring effects. On the other hand, D-BPF consistently produces clear and detailed reconstructions regardless of the number of projections. Subjectively, even with $N = 251$, D-BPF achieves image quality comparable to V-FBP with $N = 1001$. However, as $N$ reaches 2001, the reconstructed image from V-FBP appears sharper.

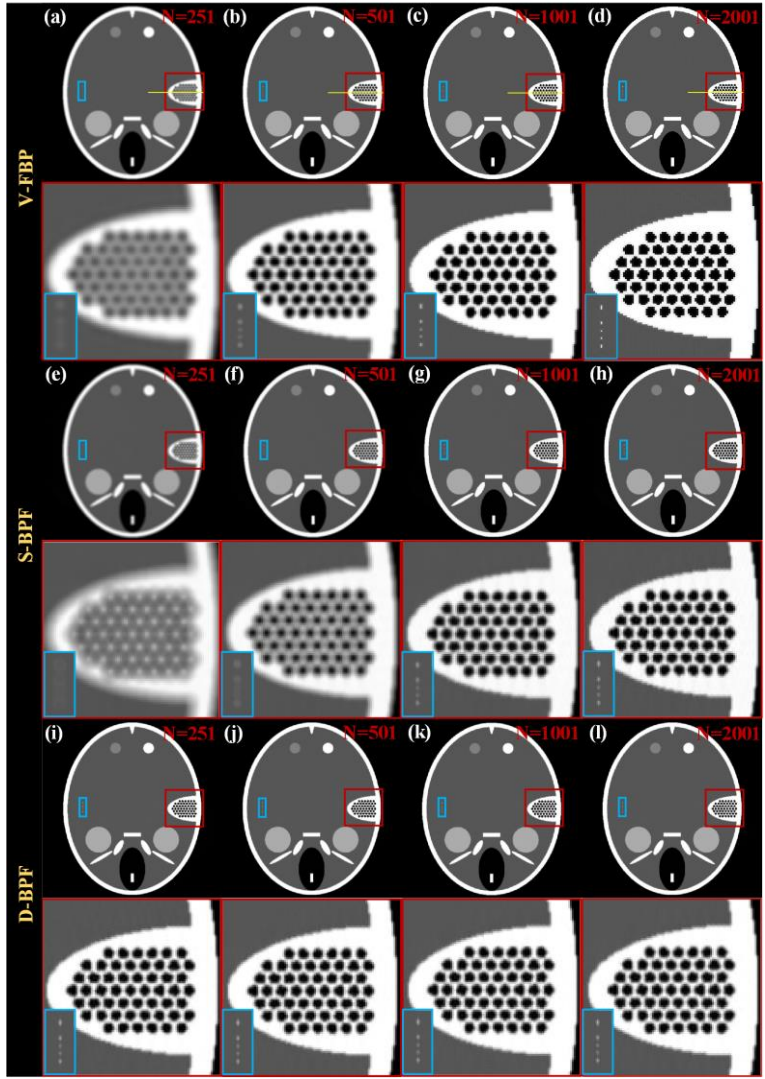

**Fig. 7** Reconstruction results via V-FBP, S-BPF and D-BPF. (a)-(d) Results of V-FBP from $N$=251 to 2001. (e)-(h) Results of S-BPF. (i)-(l) Results of D-BPF. The reconstructed images consist of 512 × 512 pixels with pixel size being 16.5 μm$^2$. The display window is [0, 3].

Fig. 8 displays the horizontal central profiles of pixels 290 to 460 from the reconstructed images. These profiles capture high-frequency components that correspond to edges and details within the reconstructed images. As the number of projections per STCT ($N$) increases, the high-frequency components of the profiles obtained from V-FBP and S-BPF gradually become more pronounced, indicating richer details in the reconstructions. In contrast, the profiles of D-BPF remain closer to those of the phantom. When *N* is below 1001, the profiles of D-BPF clearly outperform the others, consistent with the observed trend in Fig. 7. However, when *N* is increased to 2001 (Fig. 8(d)), the profiles of V-FBP and S-BPF also approach those of the phantom, albeit V-FBP being notably higher.

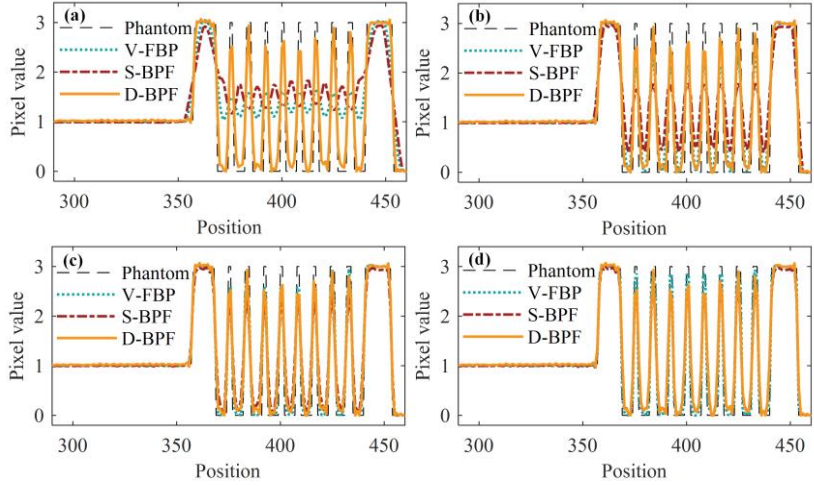

**Fig. 8** The horizontal central profiles of pixels 290 to 460 of the reconstructed images via V-FBP, S-BPF and D-BPF. (a)–(d) Profiles of the reconstructed images when the number *N* of projections per STCT is 251, 501, 1001, 2001, respectively.

Table 3 provides a summary of the quantitative metrics. The trend observed aligns with the aforementioned observations: as the number of projections (*N*) increases, the image quality improves gradually for V-FBP and S-BPF. When *N* is below 1001, D-BPF exhibits superior image quality compared to V-FBP and S-BPF under the same conditions. As *N* reaches 1001, V-FBP shows a slight advantage over the others, although the difference with D-BPF is minimal. However, when *N* reaches 2001, V-FBP clearly outperforms both S-BPF and D-BPF.

**Table 3.** Quantitative evaluation result of the reconstruction images.

| N | V-FBP | | | S-BPF | | | D-BPF | | |
|---|---|---|---|---|---|---|---|---|---|
| | RMSE | PSNR | FSIM | RMSE | PSNR | FSIM | RMSE | PSNR | FSIM |
| 251 | 0.2502 | 21.5780 | 0.8682 | 0.3126 | 19.6425 | 0.8048 | 0.1384 | 26.7212 | 0.9734 |
| 501 | 0.1712 | 24.8710 | 0.9574 | 0.2250 | 22.4979 | 0.9149 | 0.1382 | 26.7352 | 0.9740 |
| 1001 | 0.1061 | 29.0278 | 0.9838 | 0.1619 | 25.3559 | 0.9617 | 0.1381 | 26.7352 | 0.9740 |
| 2001 | 0.0446 | 36.5546 | 0.9924 | 0.1411 | 26.5511 | 0.9743 | 0.1381 | 26.7352 | 0.9740 |

*5.1.2 3D BPF reconstructions for mSTCT*

Figs. 9(a)-(c) and 9(d)-(f) illustrate the center slices in three orthogonal directions of reconstructed phantoms using V-FDK, 3D S-BPF, and 3D D-BPF, with the number of projections per STCT (*N*) set to 251 and 2001, respectively. When *N* is 251, V-FDK exhibits ring artifacts, while 3D D-BPF produces the highest quality with a clear image boundary, and 3D S-BPF appears comparatively fuzzy. As *N* increases to 2001, all three algorithms yield high-quality reconstructions. However, due to the reduced presence of redundant data away from the central plane, all three algorithms exhibit fewer artifacts, especially considering the cone-angle reaching up to 22.6°.

Additionally, the proposed 3D BPF algorithms introduce some additional artifacts in the off-center transverse slices due to the virtual Hilbert filter lines and half-scan node of mSTCT, but these minor artifacts are generally acceptable in practical applications [8].

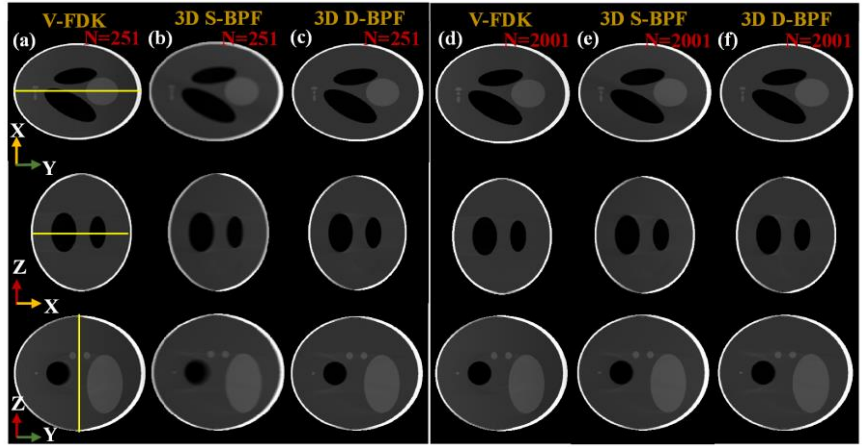

**Fig. 9** 3D reconstruction results via V-FDK, 3D S-BPF and 3D D-BPF. (a)-(c) show three central slices in three orthogonal directions when $N$ = 251, respectively. The display window is [0, 1].

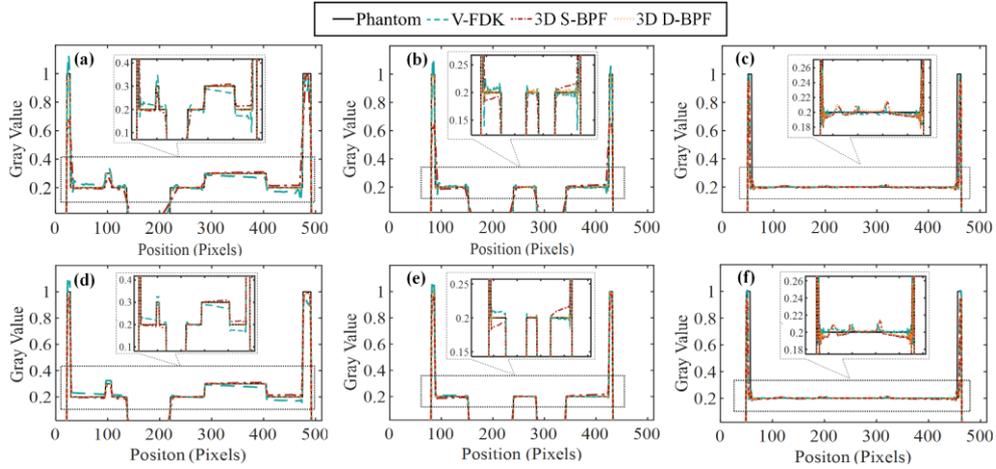

**Fig. 10** The three orthogonal central profiles of the reconstructed 3D volumes via V-FDK, 3D S-BPF and 3D D-BPF. (a) - (c) illustrate three orthogonal center profiles when the number $N$ of projections per STCT equals 251, respectively. (d) - (f) illustrate three orthogonal center profiles when $N$ = 2001.

**Table 4.** Quantitative evaluation result of the 3D reconstruction images.

| $N$ | V-FDK | | | 3D S-BPF | | | 3D D-BPF | | |
|---|---|---|---|---|---|---|---|---|---|
| | RMSE | PSNR | FSIM | RMSE | PSNR | FSIM | RMSE | PSNR | FSIM |
| 251 | 0.0557 | 27.7974 | 0.9164 | 0.0740 | 25.1892 | 0.9072 | 0.0488 | 28.9738 | 0.9465 |
| 501 | 0.0499 | 28.7520 | 0.9249 | 0.0614 | 26.8115 | 0.9283 | 0.0488 | 28.9772 | 0.9493 |
| 1001 | 0.0458 | 29.5023 | 0.9452 | 0.0543 | 27.8832 | 0.9341 | 0.0488 | 28.9777 | 0.9498 |
| 2001 | 0.0445 | 29.7463 | 0.9658 | 0.0522 | 28.2246 | 0.9352 | 0.0488 | 28.9777 | 0.9499 |

Figs. 10(a)-(c) and 10(d)-(f) display profiles in the three central orthogonal X-Y, X-Z, and Y-Z slices (indicated by yellow lines in Fig. 9) for $N$ values of 251 and 2001, respectively. In the central horizontal profiles of the X-Y and X-Z slices, a shift phenomenon in gray values can be observed. This shift is attributed to the non-homogeneous ramp-filter used in V-FDK, which hinders accurate

reconstruction in half-scans consisting of 5 STCTs [38]. As a result, 3D S-BPF also exhibits this shift, while 3D D-BPF effectively avoids this issue. Table 4 presents the three metrics, yielding results consistent with the aforementioned 2D simulated experiments. When *N* is below 1001, 3D D-BPF outperforms V-FDK and 3D S-BPF, but V-FDK surpasses the others when *N* reaches 2001.

*5.2 Results of real data*

*5.2.1 2D BPF reconstructions for mSTCT*

Fig. 11 showcases 2D reconstructed images and their corresponding zoom-in versions obtained using V-FBP, S-BPF, and D-BPF with varying numbers of realistic projections. The results affirm the effectiveness of our proposed 2D BPF reconstructions on realistic projections. The reconstructed images have dimensions of 2000 × 2000 pixels, with each pixel size measuring 5 × 5 μm².

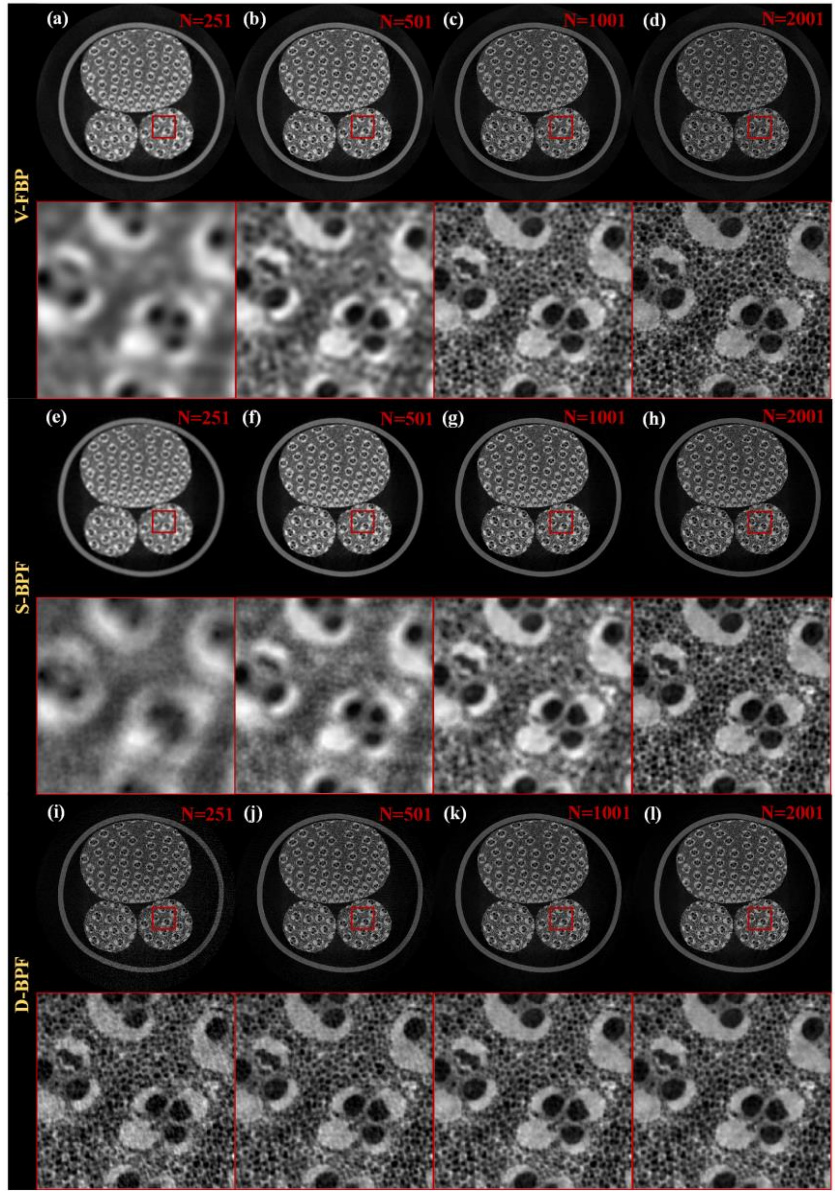

**Fig. 11** 2D reconstructed images and corresponding zoom-in images via V-FBP, S-BPF and D-BPF with different numbers of realistic projections. (a)-(d) Results of V-FBP. (e)-(h) and (i)-(l) Results of S-BPF and D-BPF. The display window is the same.

As the number of projections increases, the reconstructed images become sharper, revealing more intricate details, especially in the form of tiny holes. D-BPF exhibits superior performance when $N$ is below 1001. Even with only 251 projections, D-BPF clearly depicts the characteristics of various tiny holes in the reconstructed zoomed-in images. Combining these results with the simulated experiments mentioned earlier, we can conclude that D-BPF can save approximately 3750 projections (i.e., $(1001 - 251) \times 5 = 3750$), compared to V-FBP with $N$ set to 1001. This means that our proposed D-BPF technique can save about 75% of the projections (calculated as $3750 / (1001 \times 5) \times 100\% = 75\%$). As $N$ increases to 2001, all three algorithms capture details in the zoomed-in images, with V-FBP outperforming the others.

*5.2.2 3D BPF reconstructions for mSTCT*

Fig. 12 displays the 3D reconstructed rendered volumes and their corresponding zoomed-in regions obtained using V-FDK, 3D S-BPF, and 3D D-BPF. The reconstructed data successfully captures the structures and details of the entire bamboo, confirming the effectiveness of our proposed 3D BPF algorithms for reconstructing real 3D data. Additionally, we observe that the 3D reconstruction results align closely with the 2D experimental findings (see Fig. 11). This indicates that for high-resolution image reconstruction, 3D D-BPF is more practical for fast imaging due to its ability to save approximately 75% of the projections. It is worth mentioning that this paper only incorporates air correction for realistic projections, and further correction techniques will be employed in future studies to enhance the image quality.

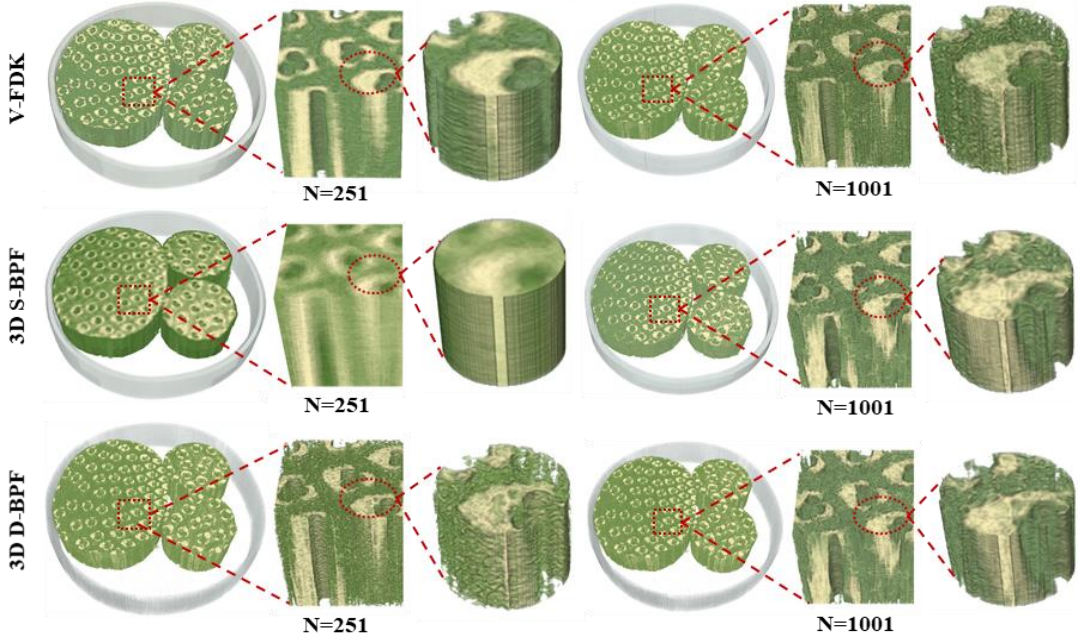

**Fig. 12** Reconstructed and rendered 3D bamboo sample volumes via V-FDK, S-BPF, and D-BPF with different numbers of real projections. The 3D complete image size is 2000 × 2000 × 256 voxels, with a voxel size of 5 × 5 × 5 μm3. The sizes of the 3D cubic region of interest in the middle and the columns is 200 × 200 × 200 voxels and 100 × 100 × 100 voxels, respectively.

## 6. Discussions

In this paper, the proposed two types of BPF reconstruction algorithms, including D-BPF and S-BPF, can directly and successfully reconstruct images, especially D-BPF, which produces greater image quality with fewer projections than others. The implemented difference between D-BPF and S-BPF mainly lies in different derivative directions, whereas two formulae can produce different resolutions in reconstruction images. The main reason for this advantage of D-BPF is that its derivative (the derivative is equivalent to the filtering process) is along the detector row, while the filtering of the proposed S-BPF and the compared V-FBP (it is V-FDK in 3D reconstruction) are filtered along the source trajectory. In D-DBP, the cut-frequency of the filtering is determined by the detector sampling interval, while it depends on the source sampling interval in S-DBP. Since the detector pixel resolution is typically much higher than the source translation steps, higher-resolution images can be obtained from D-BPF than those from S-BPF and V-FBP. Fig. 13(a) shows the sampling geometry of V-FBP (it is also a top view of V-FDK in 3D reconstruction), whose spatial resolution formulae are $\Delta\delta = \Delta\lambda \cdot h/(l+h)$ and $\Delta\lambda = 2s/(N-1)$. Based on these formulae, we can infer that, to achieve the reconstructed volume with higher spatial resolution (smaller $\Delta\delta$), the smaller sampling interval $\Delta\lambda$ and the bigger number of projections $N$ per STCT are required. In mSTCT, the total projections required will be $T \cdot N$, which means a lower scanning and reconstruction efficiency. In addition, the increase of $N$ will not increase the resolution infinitely, because the projections will not contain details smaller than $B_w/2$ ($B_w$ is the equivalent beam width).

Fig. 13(b) describes the 2D sampling geometry of D-BPF and S-BPF, and its spatial resolution formula is $\Delta\delta = \Delta u \cdot l/(l+h)$. The detector unit is typically dozens of micrometers to 200 micrometers in size. Therefore, even if the number of projections is small, D-BPF can reconstruct high-resolution images. Rays converging on the same detector unit in V-FBP cover the whole object, which solves the truncation problem, but the sampling interval will affect the resolution. Although the sampling geometry of S-BPF is consistent with that of D-BPF, its projections are differentiated along the source trajectory. In fact, the differential is a special ramp filter with phase shift and amplitude amplification. A large $\Delta\delta$ results in a small cut-off frequency ($f = 1/(2\Delta\delta)$) in the Fourier space, and hence high-frequency information is filtered out.

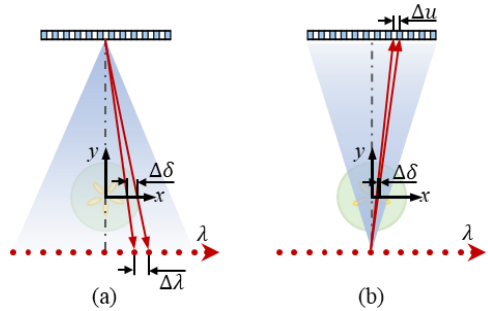

**Fig. 13** Schematic diagrams showing the 2D sampling geometry of V-FBP (a) and BPF reconstructions (b).

In fact, D-BPF still has some limitations. Though it can use fewer projection data to reconstruct a high-resolution image, its image uniformity deteriorates as the number of projections decreases. Sufficient intensive sampling points are required to obtain the satisfied image uniformity. Too few projections will introduce sparse artifacts. Besides, the parameter *L* in the backprojection weighting factor $1/L^2$ denotes the distance from the source to the reconstructed point. According to the requirement for large geometric magnification to achieve high resolution (i.e., the measured object is close to the source) and the fact that mSTCT is actually a short scan mode, the factor $1/L^2$ is unstable, and there are some reconstructed errors (i.e., the values of the reconstructed regions near the source that have not been measured twice are large, manifested in the shift phenomenon in the profiles and nonuniform resolution in the reconstructed images) [38]. Compared to D-BPF, although S-BPF requires a larger number of projections, it also exhibits some advantages. On the one hand, its backprojection weighting factor is $1/H^2$. Since $H$ is always larger than $L$ for high-resolution imaging, S-BPF will obtain more stable results than D-BPF when noise is present. On the other hand, the projection is not truncated along the source direction but is truncated along the detector direction in STCT. As a result, unlike D-BPF, S-BPF can entirely prevent the introduction of errors resulting from truncated projections. Moreover, both 3D D-BPF and 3D S-BPF show some artifacts at positions off the central plane (see the X-Z and Y-Z slices in Fig. 10), which are caused by a little redundancy and the half-scan mode. To reduce these artifacts, we are attempting to implement a new weighting strategy in the full-scan way. Finally, D-BPF and S-BPF introduce multiple rotations of the coordinate in the finite Hilbert inverse (see Fig. 3 and Fig. 5), and the rotation can be achieved by the interpolation, which can smooth a part of the high-frequency information. To further improve the reconstructed quality, we are studying how to complete the finite Hilbert inverse function in arbitrary directions by utilizing deep learning technology without any interpolation operations.

## 7. Conclusion

In this paper, we propose two types of BPF reconstruction algorithms for mSTCT imaging geometry, including S-BPF and D-BPF, and present the cases in 2D and 3D reconstructions. In the proposed algorithms, to decrease the truncation artifacts and reduce the number of projections, we derive the DBP formulae for STCT from different filtering directions, including the differential along the source trajectory and the detector (i.e., the S-DBP and D-DBP formulae, respectively), design the finite Hilbert inverse function per STCT, and simplify the DBP formulae for STCT. Results of massive simulated and real experiments demonstrate that the proposed BPF algorithms for mSTCT can directly and effectively complete the reconstruction with truncated projections in mSTCT. Compared with the previous methods, D-BPF can reconstruct a high-resolution image with fewer projections, and our results report that the number of projections can be reduced by 3750 via D-BPF reconstruction, which improves the efficiency of data acquisition and reduces the high requirements for storage and computation.

**Funding Information**

This research is supported by the National Natural Science Foundation of China (No. 52075133, 62171067), the National Key Research and Development Plan of China (No. 2022YFF0706400), the CGN-HIT Advanced Nuclear and New Energy ResearchInstitute (No. CGN-HIT202215), and the China Scholarship Council under Grant (No. 202006050115).

**Declaration of Competing Interest**

The authors declare that they have no conflict of interest.

**Data Availability**

Data will be made available on request.

**Acknowledgments**

The authors would like to thank Prof. Hengyong Yu for his valuable discussions and constructive suggestions.

The manuscript is approved by all authors for publication. We would like to declare that the work described was original research that has not been published previously, and not under consideration for publication elsewhere, in whole or in part.

**Supplementary**

**A. Calculation of coordinate transformation from parallel-beam to STCT fan-beam geometry**

According to the geometry of a STCT depicted in Fig. 1(c), the relationships between variables of parallel-beam to fan-beam geometry are derived as follows:

$$r = \frac{\lambda h + ul}{\sqrt{(l+h)^2 + (\lambda-u)^2}}, \varphi = \theta + \arctan(\frac{\lambda-u}{l+h}), \tag{A.1}$$

to realize coordinate transformation, a Jacobian determinant $|J|$ will be introduced, i.e.,

$$\mathrm{d}r\mathrm{d}\varphi = |J|\mathrm{d}\lambda\mathrm{d}u = \begin{vmatrix} \frac{\partial \varphi}{\partial u} & \frac{\partial \varphi}{\partial \lambda} \\ \frac{\partial r}{\partial u} & \frac{\partial r}{\partial \lambda} \end{vmatrix} \mathrm{d}\lambda\mathrm{d}u = \frac{(l+h)^2}{\left((l+h)^2+(\lambda-u)^2\right)^{\frac{3}{2}}}\mathrm{d}\lambda\mathrm{d}u. \tag{A.2}$$

Here are the differential formulae for each item:

$$\frac{\partial \varphi}{\partial \lambda} = \frac{1}{1+\left(\frac{\lambda-u}{l+h}\right)^2} \times \frac{1}{l+h} = \frac{(l+h)}{(l+h)^2+(\lambda-u)^2}, \frac{\partial \varphi}{\partial u} = \frac{1}{1+\left(\frac{\lambda-u}{l+h}\right)^2} \times \frac{-1}{l+h} = \frac{-(l+h)}{(l+h)^2+(\lambda-u)^2},$$

$$\frac{\partial r}{\partial \lambda} = \frac{h}{\sqrt{(l+h)^2+(\lambda-u)^2}} - \frac{(\lambda-u)(\lambda h+ul)}{\left((l+h)^2+(\lambda-u)^2\right)^{\frac{3}{2}}}, \frac{\partial r}{\partial u} = \frac{l}{\sqrt{(l+h)^2+(\lambda-u)^2}} + \frac{(\lambda-u)(\lambda h+ul)}{\left((l+h)^2+(\lambda-u)^2\right)^{\frac{3}{2}}}.$$

$$\tag{A.3}$$

Additionally, the relationships between the sine and cosine of the fan-angle are

$$\cos(\varphi-\theta) = \frac{l+h}{\sqrt{(l+h)^2+(\lambda-u)^2}}, \sin(\varphi-\theta) = \frac{\lambda-u}{\sqrt{(l+h)^2+(\lambda-u)^2}}, \tag{A.4}$$

by introducing the sum-to-product identities, we get

$$\cos\varphi = \frac{(l+h)\cos\theta - (\lambda-u)\sin\theta}{\sqrt{(l+h)^2+(\lambda-u)^2}}, \sin\varphi = \frac{(l+h)\sin\theta + (\lambda-u)\cos\theta}{\sqrt{(l+h)^2+(\lambda-u)^2}}. \tag{A.5}$$

Finally, we can obtain

$$r = x\cos\varphi + y\sin\varphi = \frac{(l+h)(x\cos\theta + y\sin\theta) - (\lambda-u)(x\sin\theta - y\cos\theta)}{\sqrt{(l+h)^2+(\lambda-u)^2}}, \tag{A.6}$$

therefore,

$$r-r' = \frac{(l+h)(x\cos\theta+y\sin\theta) - u(-x\sin\theta+y\cos\theta+l) - \lambda(x\sin\theta-y\cos\theta+h)}{\sqrt{(l+h)^2+(\lambda-u)^2}}, \tag{A.7}$$

let $H = x\sin\theta - y\cos\theta + h$ and $L = -x\sin\theta + y\cos\theta + l$ in this case; Eq. (A.7) can be simplified as follows:

$$r - r' = \frac{(l+h)(x\cos\theta + y\sin\theta) - uL - \lambda H}{\sqrt{(l+h)^2 + (\lambda - u)^2}}. \tag{A.8}$$

**B. Calculating and simplifying the D-DBP formula**

For the D-DBP formula in Eq. (16) filtered along the source translation,

$$Db_\eta^\theta(\vec{x}) = \frac{1}{2} \int_{-\infty}^{+\infty} \frac{1}{L^2} \frac{\partial}{\partial u} \left\{ \text{sgn}\left[\sin\left(\theta + \arctan\left(\frac{\lambda - u}{l+h}\right) - \eta\right)\right] \frac{(l+h)^2}{\sqrt{(l+h)^2 + (\lambda - u)^2}} p_\theta(\lambda, u) \right\} \bigg|_{u=u^*} d\lambda, \tag{B.1}$$

we apply the product rule to yield

$$Db_\eta^\theta(\vec{x}) = \frac{1}{2} \int_{-\infty}^{+\infty} \frac{1}{L^2} \text{sgn}\left[\sin\left(\theta + \arctan\left(\frac{\lambda - u^*}{l+h}\right) - \eta\right)\right] \frac{\partial}{\partial u} \left\{ \frac{(l+h)^2}{\sqrt{(l+h)^2 + (\lambda - u)^2}} p_\theta(\lambda, u) \right\} \bigg|_{u=u^*} d\lambda$$

$$+ \frac{1}{2} \int_{-\infty}^{+\infty} \frac{1}{L^2} \frac{(l+h)^2}{\sqrt{(l+h)^2 + (\lambda - u^*)^2}} p_\theta(\lambda, u^*) \frac{\partial}{\partial u} \left\{ \text{sgn}\left[\sin\left(\theta + \arctan\left(\frac{\lambda - u}{l+h}\right) - \eta\right)\right] \right\} \bigg|_{u=u^*} d\lambda. \tag{B.2}$$

For the second term, note that $u^*$ is a function of $\lambda$ (see Eq. (12)), and the derivative of the signum function $\text{sgn}(x)$ is known to be the Dirac delta function $\delta(x)$. By applying the chain rule, it is written as

$$\text{S.T.} = \frac{1}{2} \int_{-\infty}^{+\infty} \frac{1}{L^2} \frac{(l+h)^3 p_\theta(\lambda, u^*)}{\left((l+h)^2 + (\lambda - u^*)^2\right)^{\frac{3}{2}}} \delta\left(\sin\left(\theta + \arctan\left(\frac{\lambda - u^*}{l+h}\right) - \eta\right)\right) \cos\left(\theta + \arctan\left(\frac{\lambda - u^*}{l+h}\right) - \eta\right) d\lambda. \tag{B.3}$$

Assuming that parameters $l$, $h$ are known and integration variables $\lambda \in (-\infty, \infty), u \in (-\infty, \infty)$, and the angles $\theta$, $\eta$ that respectively reflect the source translation and the Hilbert filtering direction are the period of $2\pi$. According to the properties of the Dirac delta function $\delta(x)$, some solution sets of $\lambda$ and $u^*$ can be obtained by solving roots of the equation $\sin\left(\theta + \arctan\left(\frac{\lambda - u^*}{l+h}\right) - \eta\right) = 0$. Specifically, because the arctan function is monotonically increasing, there is only one set of solutions to $\lambda$ and $u^*$ for this function.

1) When $-\pi/2 < \theta - \eta < \pi/2$, there is a set of solutions as follows:

$$u_1^* = \lambda_1 + (l+h)\tan(\theta - \eta), \tag{B.4}$$

where $u_1^* = \frac{(l+h)(x\cos\theta + y\sin\theta) - \lambda_1 H}{L}$ (see Eq. (12)).

According to composite function $\delta(g(x))$ and its integral characteristics,

$$\int_{-\infty}^{\infty} f(x)\delta(g(x))dx = \sum_{k=1}^{N} \frac{f(x_k)}{|g'(x_k)|}, \tag{B.5}$$

Eq. (B.3) can be summarized as follows:

$$\text{S.T.} = \frac{1}{2L^2} \frac{(l+h)^3 p_\theta(\lambda_1, u_1^*)}{\left((l+h)^2 + (\lambda_1 - u_1^*)^2\right)^{\frac{3}{2}}} \frac{\cos\left(\theta + \arctan\left(\frac{\lambda_1 - u_1^*}{l+h}\right) - \eta\right)}{\left|\frac{\partial}{\partial \lambda} \sin\left(\theta + \arctan\left(\frac{\lambda_1 - u_1^*}{l+h}\right) - \eta\right)\right|_{\lambda=\lambda_1}}$$
$$= \frac{1}{2L^2} \frac{(l+h)^2 p_\theta(\lambda_1, u_1^*)}{\sqrt{(l+h)^2 + (\lambda_1 - u_1^*)^2}} \frac{\text{sgn}\left(\cos\left(\theta + \arctan\left(\frac{\lambda_1 - u_1^*}{l+h}\right) - \eta\right)\right)}{\left|1 - \left(\frac{\partial u^*}{\partial \lambda}\right)_{\lambda=\lambda_1}\right|}, \quad \text{(B.6)}$$

with

$$\left(\frac{\partial u^*}{\partial \lambda}\right)_{\lambda=\lambda_1} = \left(\frac{\partial}{\partial \lambda} \frac{(l+h)(x\cos\theta + y\sin\theta) - \lambda H}{L}\right)_{\lambda=\lambda_1} = -\frac{H}{L}. \quad \text{(B.7)}$$

Note that $\cos\left(\theta + \arctan\left(\frac{\lambda_1 - u_1^*}{l+h}\right) - \eta\right) = \cos 0 = 1$, Eq. (B.6) can be simplified as

$$\text{S.T.} = \frac{1}{2|L|} \frac{(l+h) p_\theta(\lambda_1, u_1^*)}{\sqrt{(l+h)^2 + (\lambda_1 - u_1^*)^2}}, \quad \text{(B.8)}$$

where $u_1^* = x\cos\theta + y\sin\theta - H\tan(\theta - \eta)$ and $\lambda_1 = x\cos\theta + y\sin\theta - L\tan(\theta - \eta)$, by solving the Eq. (B.4).

For special cases, such as $\theta - \eta = 0$, the solution set relationship satisfies: $u_1^* = \lambda_1 = x\cos\theta + y\sin\theta$. In this case, the direction of Hilbert transform is vertical to the direction of the source translation trajectory.

2) When $\theta - \eta \in (-\infty, -\pi/2] \cup [\pi/2, \infty)$, the equation $\sin\left(\theta + \arctan\left(\frac{\lambda - u^*}{l+h}\right) - \eta\right) = 0$ has no solution set about $u^*$ and $\lambda$, so the value of the second term **S.T. is equal to 0.** For special cases, such as $\theta - \eta = \pm\pi/2$, the direction of Hilbert transform is parallel to the trajectory of source translation. Further, only if $\theta - \eta = \pi/2$, the direction of Hilbert transform is consistent with the direction of the source translation trajectory, and we get $\pi > \theta + \arctan\left(\frac{\lambda - u^*}{l+h}\right) - \eta > 0$, thus $\text{sgn}\left[\sin\left(\theta + \arctan\left(\frac{\lambda - u^*}{l+h}\right) - \eta\right)\right] \equiv 1$.